%% file: 00_main.tex
\documentclass[11pt,a4paper]{article}
\usepackage[hyperref]{eacl2021}
\usepackage{times}
\usepackage{latexsym}

\usepackage{pifont}% http://ctan.org/pkg/pifont
\usepackage{balance}
\usepackage{helvet}  %Required
\usepackage{courier}  %Required
\usepackage{graphicx}  %Required
\usepackage{amssymb}
\usepackage{amsmath}
\usepackage{algorithm}
\usepackage{algorithmic}
\usepackage{amsthm}
\usepackage{multirow}
\usepackage{pgffor}
\usepackage{graphicx}
\usepackage{epstopdf}
\usepackage{tikz}
\usepackage{epstopdf}
\usepackage{mathtools}
\usepackage{enumitem}
\usepackage{subfigure}
\usepackage{tabulary}
\usepackage{color}
\usepackage{url}
\usepackage{flushend}
\usepackage{stfloats}
\usepackage{tcolorbox}
\usepackage[utf8]{inputenc}
\usepackage[english]{babel}
\usepackage{tabularx}
\usepackage{CJKutf8}
\usepackage{microtype}
\usepackage{textcomp}
\usepackage{booktabs}
\usepackage{colortbl}
\usepackage{soul}
\usepackage{bm}
\usepackage{xcolor}
\usepackage{appendix}
\usepackage{xspace}
\usepackage{microtype}
\newcommand{\snat}{\textbf{SNAT}\xspace}

\aclfinalcopy % Uncomment this line for the final submission
\newcount\DraftStatus  % 0 suppresses notes to selves in text
\usepackage{color}
\definecolor{darkgreen}{rgb}{0,0.5,0}
\definecolor{purple}{rgb}{1,0,1}
\definecolor{todocolor}{rgb}{0.9,0.1,0.1}
\definecolor{hycolor}{rgb}{0.7,0.7,0.3}
\definecolor{fixcolor}{rgb}{0.1,0.7,0.3}

\newcommand{\draftnote}[2]{\ifnum\DraftStatus=1
	\marginpar{
		\tiny\raggedright
		\hbadness=10000
		\def\baselinestretch{0.8}
		\textcolor{#1}{\textsf{\hspace{0pt}#2}}}
	\fi}

\title{Syntactic and Semantic-Augmented Non-Autoregressive \\Transformer for Neural Machine Translation}
\title{Syntactic and Semantic Structure Augmented Non-Autoregressive Transformer for Neural Machine Translation} %
\title{Enriching Non-Autoregressive Transformer with Syntactic and Semantic Structures for Neural Machine Translation}

\author{Ye Liu$^1$, Yao Wan$^2$, Jian-Guo Zhang$^1$, Wenting Zhao$^1$, Philip S. Yu$^{1}$\\
\normalsize$^1$Department of Computer Science, University of Illinois at Chicago, Chicago, IL, USA\\
\normalsize$^2$School of Computer Science and Technology, Huazhong University of Science and Technology, Wuhan, China\\
\{yliu279, jzhan51, wzhao41, psyu\}@uic.edu, wanyao@hust.edu.cn
}

\begin{document}
\maketitle
\begin{abstract}
% Most neural machine translation systems (NMT) using autoregressive models (AT) which generate text autoregressively from left to right achieve state-of-the-art performance. 
% However, due to the unparallelizable nature of the autoregressive factorization, these models suffer from heavy latency during inference. Recently, non-autoregressive models (NAT) have been proposed to decrease the inference time.
% However, the decoding process of each token is conditionally independent of others in those models. Such a generation process tend to make the output sentence inconsistent, and thus the learned non-autoregressive models only achieve inferior accuracy compared to their autoregressive counterparts. 
% Yao Wan, feel free to modify it.
% In most neural machine translation (NMT) systems, autoregressive models that generate text from left to right have achieved decent performance. 
% However, due to the unparallelizable nature of the autoregressive factorization, these models suffer from heavy latency during inference. 
% Recently, non-autoregressive models have been proposed to decrease the inference time. 
% Despite the gained efficiency of non-autoregressive models, such a generation just achieves inferior performance compared with the autoregressive models. 
The non-autoregressive models have boosted the efficiency of neural machine translation through parallelized decoding at the cost of effectiveness, when comparing with the autoregressive counterparts.
% We state that incorporating structured information which can provide rich syntactic and semantic information can improve the performance of NAT. 
In this paper, we claim that the syntactic and semantic structures among natural language are critical for non-autoregressive machine translation and can further improve the performance. However, these structures are rarely considered in the existing non-autoregressive models.
% In this paper, we propose a novel \textbf{S}tructure-aware \textbf{N}on-\textbf{A}utoregressive Machine \textbf{T}ranslation (\textsc{SNAT}), which proposes to incorporate explicit structure information in transformer and consider intermediate latent alignment with the target to learn the long-term token dependencies.
% Based on this intuition, we propose a novel syntactic and semantic structure-aware non-autoregressive approach for machine translation by incorporating the explicit syntactic and semantic structure into a Transformer.
Inspired by this intuition, we propose to incorporate the explicit syntactic and semantic structures of languages into a non-autoregressive Transformer, for the task of neural machine translation.
% , which proposes to incorporate explicit structure information in transformer and consider intermediate latent alignment with the target to learn the long-term token dependencies.
Moreover, we also consider the intermediate latent alignment within target sentences to better learn the long-term token dependencies.
% We conduct experiments on two real-world datasets (i.e., WMT14 En-De and WMT16 En-Ro) to verify the efficiency and effectiveness of our proposed model. 
% Experimental results show that our model achieves a significantly faster speed, as well as keeps the translation quality when compared with several state-of-the-art non-autoregressive models.
Experimental results on two real-world datasets (i.e., WMT14 En-De and WMT16 En-Ro) show that our model achieves a significantly faster speed, as well as keeps the translation quality when compared with several state-of-the-art non-autoregressive models.

\end{abstract}
% process tends to make the output sentence inconsistent, and thus

\input{01_Intro}
% \input{02_REL}
\input{03_MTD}

\input{04_EXP}

\section{Conclusion}
In this paper, we have proposed a novel syntactic and semantic structure-aware non-autoregressive Transformer model \snat for NMT. 
% The proposed \snat model considers both syntactic and semantic structure information existing in natural language, greatly reducing the time cost for inference compared with autoregressive Transformer models and most NAT models. 
The proposed model aims at reducing the computational cost in inference as well as keeping the quality of translation by incorporating both syntactic and semantic structures existing among natural languages into a non-autoregressive Transformer. In addition, we have also designed an intermediate latent alignment regularization within target sentences to better learn the long-term token dependencies.
% Experiments show that our method gains significant performance improvements against existing state-of-the-art NAT models and shows competitive results with autoregressive Transformer models.
% Comprehensive experiments and analysis on two real-world datasets (i.e., WMT14 En$\to$De and WMT16 En$\to$Ro) verify the efficiency and effectiveness of our proposed approach when compared with existing state-of-the-art autoregressive and non-autoregressive Transformer models.
Comprehensive experiments and analysis on two real-world datasets (i.e., WMT14 En$\to$De and WMT16 En$\to$Ro) verify the efficiency and effectiveness of our proposed approach. %when compared with existing state-of-the-art autoregressive and non-autoregressive Transformer models.

\section*{Acknowledgements}
This work is supported in part by NSF under grants III-1763325, III-1909323, and SaTC-1930941. 

% The acknowledgements should go immediately before the references.  Do

% \bibliography{05_REF}
% \bibliographystyle{acl_natbib}

\input{output.bbl}
\end{document}

%% file: 01_Intro.tex
\section{Introduction}
% Parallelization is an important ingredient to make deep learning models computationally tractable. Most recent state-of-the-art models in neural machine translation are autoregressive~\cite{wu2016google,ott2018scaling}, while autoregressive decoders requiring sequential execution cannot be parallelized in the inference stage.
% Recently, non-autoregressive neural machine translation based on Transformer~(NAT)~\cite{gu2018non}, which aims to enable the parallel generation of output tokens without sacrificing translation quality, has attracted much attention.
Recently, non-autoregressive models~\cite{gu2018non}, which aim to enable the parallel generation of output tokens without sacrificing translation quality, have attracted much attention. 
Although the non-autoregressive models have greatly sped up the inference process for real-time neural machine translation (NMT)~\cite{gu2018non}, their performance is considerably worse than that of autoregressive counterparts. Most previous works attribute the poor performance to the inevitable conditional independence issue when predicting target tokens, and many variants have been proposed to solve it. 
For example, several techniques in non-autoregressive models are investigated to mitigate the trade-off between speedup and performance, including iterative refinement~\cite{lee2018deterministic}, insertion-based models~\cite{chan2019kermit,stern2019insertion}, latent variable based models \cite{kaiser2018fast,shu2020latent}, CTC models~\cite{libovicky2018end,saharia2020non}, alternative loss function based models~\cite{wei2019imitation,wang2019non,shao2020minimizing}, and masked language models~\cite{ghazvininejad2019mask,ghazvininejad2020aligned}.

% \newcite{lee2018deterministic} and \newcite{ghazvininejad2019mask} proposed an inference strategy with an adaptive number of steps to minimize the generation latency without sacrificing the generation quality. 
% There are also some other works trying to use alternative loss function from token-wise to phrase-wise, such as using dynamic programming and reinforcement learning to calculate the alignment between any prefix of the target translation sentence and predicted translation sentence~\cite{wei2019imitation,li2019hint,wang2019non,shao2020minimizing,ghazvininejad2020aligned}.
% In the other works such as \cite{libovicky2018end,sun2019fast,li2020lava}, the authors modeled the local dependencies of the sentences to capture the structure among word dependencies. %between the tokens.
% Some other works \cite{kaiser2018fast,ma2019flowseq,shu2020latent} proposed to use latent space that summarizes the relevant information from the target to model the uncertainty about the target sentence.

\begin{table*}[!t]
\small
\centering
\caption{A motivating example on WMT14 En$\to$De dataset. English with POS$|$NER and its corresponding German translation with POS$|$NER. The \textcolor{blue}{Blue} labels show the same tags, while the \textcolor{red}{Red} labels show the different tags in two languages.}
\resizebox{0.9\textwidth}{!}{
\begin{tabular}{l|l}
\toprule[1.5pt]
EN:      & ~A  ~~~~republican ~ strategy ~~~~to~~~ counter~~~~ the~~~ rel-election ~~of~~~~~~~ Obama~~~~~~~~~~~.    \\
         & \small ~~$\bm{\mid}$~~~~~~~~~~~~~$\bm{\mid}$~~~~~~~~~~~~~~~~$\bm{\mid}$~~~~~~~~~~~~$\bm{\mid}$~~~~~~~~~~~$\bm{\mid}$~~~~~~~~~~~~$\bm{\mid}$~~~~~~~~~~~~~~$\bm{\mid}$~~~~~~~~~~~~~$\bm{\mid}$~~~~~~~~~~~~~~$\bm{\mid}$~~~~~~~~~~~~~~~~$\bm{\mid}$     \\
EN POS: & \small \textcolor{blue}{DET} ~~~~ \textcolor{blue}{ADJ} ~~~~~~ \textcolor{blue}{NOUN} ~~\textcolor{red}{PART}~~\textcolor{red}{VERB} ~~~\textcolor{blue}{DET} ~~~~\textcolor{blue}{NOUN}~~~ \textcolor{red}{ADP}~~~~~ \textcolor{blue}{PROPN}~~~~ \textcolor{blue}{PUNCT}\\ 
EN NER: & ~ \textcolor{blue}{O} ~~~ \textcolor{blue}{B$\_$NORP}~~~~~~~~ \textcolor{blue}{O}~~~~~~~~~ \textcolor{blue}{O}~~~~~~~~~ \textcolor{blue}{O} ~~~~~~~~~\textcolor{blue}{O} ~~~~~~~~~~~ \textcolor{blue}{O} ~~~~~~~~~~\textcolor{blue}{O} ~~~ \textcolor{blue}{B$\_$PERSON} ~~~~\textcolor{blue}{O} \\ \hline
DE:      & Eine ~~republikanische~~ strategie~~ gegen~~~~ die~~~~ wiederwahl~~~~ Obama~~~~~~~~~~~.     \\
  &\small ~~$\bm{\mid}$~~~~~~~~~~~~~~~~~$\bm{\mid}$~~~~~~~~~~~~~~~~~~~~~~~$\bm{\mid}$~~~~~~~~~~~~~~$\bm{\mid}$~~~~~~~~~~$\bm{\mid}$~~~~~~~~~~~~~~~~$\bm{\mid}$~~~~~~~~~~~~~~~~~~~$\bm{\mid}$~~~~~~~~~~~~~~~~$\bm{\mid}$     \\
DE POS: & \small \textcolor{blue}{DET} ~~~~~~~~~ \textcolor{blue}{ADJ} ~~~~~~~~~~~~~\textcolor{blue}{NOUN}~~~~\textcolor{red}{ADP}~~~~ \textcolor{blue}{DET}~~~~~~\textcolor{blue}{NOUN}~~~~~~~~ \textcolor{blue}{PROPN}~~~~ \textcolor{blue}{PUNCT}                          \\ 
EN NER: & ~ \textcolor{blue}{O} ~~~~~~~ \textcolor{blue}{B$\_$NORP}~~~~~~~~~~~~~~~ \textcolor{blue}{O}~~~~~~~~~~~ \textcolor{blue}{O}~~~~~~~~ \textcolor{blue}{O} ~~~~~~~~~~~~\textcolor{blue}{O}  ~~~~~~~~  \textcolor{blue}{B$\_$PERSON} ~~~~~\textcolor{blue}{O} \\ 
\bottomrule[1.5pt]
% EN:      & ~He~~~~~~~said~~~~~~~~~~~ he~~~~~~ then~~~ heard~~~~~ his~~~ friend~~~~~~~~~, ~~~~~~~~~hamza~~~ calling~~~~ to~~~~~~ him~~~~~~~~~ .  \\
%     & \small~~~$\bm{\mid}$~~~~~~~~~~~$\bm{\mid}$~~~~~~~~~~~~~~~~$\bm{\mid}$~~~~~~~~~~~$\bm{\mid}$~~~~~~~~~~~$\bm{\mid}$~~~~~~~~~~~$\bm{\mid}$~~~~~~~~~~~$\bm{\mid}$~~~~~~~~~~~~$\bm{\mid}$~~~~~~~~~~~~~~~$\bm{\mid}$~~~~~~~~~~~~$\bm{\mid}$~~~~~~~~~~~$\bm{\mid}$~~~~~~~~~~~$\bm{\mid}$~~~~~~~~~~~~$\bm{\mid}$     \\
% EN POS: & \small PRON~~\textcolor{blue}{VERB} ~~ PRON~~ ADV~~ VERB~~ DET~~ NOUN~~ PUNCT~~ NOUN~~ VERB~~ ADP~~ PRON~~ PUNCT     \\ \hline
% RO:      & cei mai multi dintre cei ucisi in atac erau elevi.  \\
%     & \small~~$\bm{\mid}$~~~~~~~~~~~$\bm{\mid}$~~~~~~~~~~~~~$\bm{\mid}$~~~~~~~~~~~~$\bm{\mid}$~~~~~~~~~~~$\bm{\mid}$~~~~~~~~~~~~~$\bm{\mid}$~~~~~~~~~~~~~$\bm{\mid}$~~~~~~~~~~~~~~$\bm{\mid}$~~~~~~~~~~~~$\bm{\mid}$~~~~~~~~~~~~$\bm{\mid}$~~~~~~~~~~~~~~$\bm{\mid}$~~~~~~~~~~~~~~$\bm{\mid}$~~~~~~~~~~~~~~~$\bm{\mid}$     \\
% RO POS: & \small ADV~~ VERB~~ ADV~~ NOUN~~ VERB~~ NOUN~~ PUNCT~~ ADP~~ NOUN~~ PUNCT~~ VERB~~ NOUN~~ PUNCT                       \\          \bottomrule                     
\end{tabular}
}
\label{word_pos}
\end{table*}

% These works have narrowed the performance gap between autoregressive and non-autoregressive models on machine translation. However, non-autoregressive models still suffer from syntactic and semantic limitations: the translations of non-autoregressive models contain incoherent phrases (e.g. repetitive words) and some informative tokens on the source side are absent. 
Although these works have tried to narrow the performance gap between autoregressive and non-autoregressive models, and have achieved some improvements on machine translation, the non-autoregressive models still suffer from syntactic and semantic limitations. That is, the translations of non-autoregressive models tend to contain incoherent phrases (e.g. repetitive words), and some informative tokens on the source side are absent. 
It is because that in non-autoregressive models, each token in the target sentence is generated independently. Consequently, it will cause the multimodality issue, i.e., the non-autoregressive models cannot model the multimodal distribution of target sequences properly~\cite{gu2018non}.
% Plenty of syntactic and semantic mistakes are introduced into the translated sentences due to the incoherent phrases and the omission of meaningful tokens on the source side.

One key observation to mitigate the syntactic and semantic error is that source and target translated sentences follow the same structure, which can be reflected from Part-Of-Speech (POS) tags and Named Entity Recognition (NER) labels. 
Briefly, POS, which aims to assign labels to words to indicate their categories by considering the long-distance structure of sentences, can help the model learn the syntactic structure to avoid generating the repetitive words. 
Likewise, NER, which discovers the proper nouns and verb of sentences, naturally helps the model to recognize some meaningful semantic tokens that may improve the quality of translation. This observation motivates us to leverage the syntactic as well as semantic structures of natural language to improve the performance of non-autoregressive NMT. We present a motivating example in Table~\ref{word_pos} to better illustrate our idea.
From this table, we can find that although the words are altered dramatically from the English sentence to its German translation, the corresponding POS and NER tags still remain similar. For example, most POS tags are identical and follow the same pattern, except that PART, VERB, ADP in the English do not match with the German ADP, while the NER tags are exactly the same in both sentences. 
% Thus we are motivated to enrich the NAT model with those structure information. 

In this paper, we propose an end-to-end \underline{S}yntactic and semantic structure-aware \underline{N}on-\underline{A}utoregressive \underline{T}ransformer model (\snat) for NMT. We take the structure labels and words as inputs of the model. With the guidance of extra sentence structural information, the model greatly mitigates the negative impact of the multimodality issue.
The core contributions of this paper can be summarized as that we propose 1) a syntax and semantic structure-aware Transformer which takes sequential texts and the structural labels as input and generates words conditioned on the predicted structural labels, and 
 2) an intermediate alignment regularization which aligns the intermediate decoder layer with the target to capture coarse target information.
% We conduct experiments on four benchmark tasks over two datasets, including WMT14 En-De and %WMT16 En$\leftrightarrow$Ro. 
% $\leftrightarrow$
We conduct experiments on four benchmark tasks over two datasets, including WMT14 En$\to$De and WMT16 En$\to$Ro.
Experimental results indicate that our proposed method achieves competitive results compared with existing state-of-the-art non-autoregressive and autoregressive neural machine translation models, as well as significantly reduces the decoding time.

% \begin{figure*}[t]
% \centering
% \includegraphics[width=0.95\linewidth]{table1.pdf}
% \end{figure*}
% we \textcolor{red}{xxx}.

%% file: 03_MTD.tex
\section{Background}
Regardless of its convenience and effectiveness, the autoregressive decoding methods suffer two major drawbacks. One is that they cannot generate multiple tokens simultaneously, leading to inefficient use of parallel hardware such as GPUs. The other is beam search has been found to output low-quality translation with large beam size and deteriorates when applied to larger search spaces. However, non-autoregressive transformer (NAT) could potentially address these issues. Particularly, they aim at speeding up decoding through removing the sequential dependencies within the target sentence and generating multiple target tokens in one pass, as indicated by the following equation:
\begin{equation}
\small
    P_{\rm \mathbf{NAT}}(\mathbf{y} | \mathbf{x} ; \phi)=\prod_{t=1}^{m} p\left(y_{t} | \hat{\mathbf{x}},\mathbf{x}; \phi\right), 
\end{equation}
where $\hat{\mathbf{x}} = \{\hat{x}_{1}, \ldots, \hat{x}_{m}\}$ is the copied source sentence.
% and $m$ is the length of the target sentence $\mathbf{y}^{*}$. 
Since the conditional dependencies within the target sentence ($y_{t}$ depends on $y_{<t}$) are removed from the decoder input, the decoder is unable to leverage the inherent sentence structure for prediction. Hence the decoder is supposed to figure out such target-side information by itself given the source-side information during training. This is a much more challenging task compared to the autoregressive counterparts. From our investigation, we find the NAT models fail to handle the target sentence generation well. It usually generates repetitive and semantically incoherent sentences with missing words. Therefore, strong conditional signals should be introduced as the decoder input to help the model better learn internal dependencies within a sentence.

\begin{figure*}[t]
\centering
\includegraphics[width=0.86\linewidth]{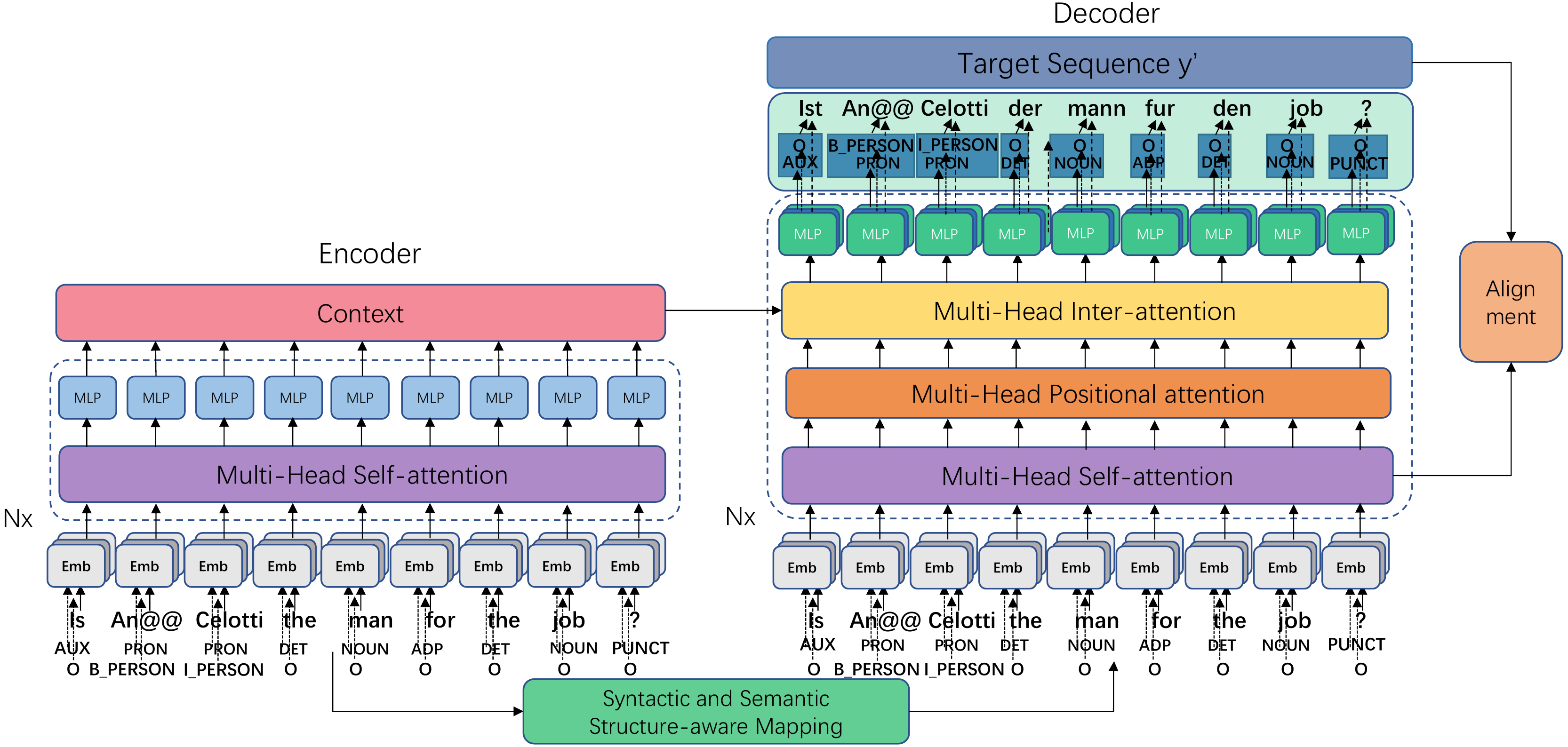}
\caption{An overview of the proposed \snat for neural machine translation.}
\label{fig:Method}
\end{figure*}

\section{Methodology} \label{sec:meth}
% In this section, we propose our model \snat to bring the structure information to the NAT model and intermediate latent space alignment with the target. 
In this section, we present our model \snat to incorporate the syntactic and semantic structure information into a NAT model as well as an intermediate latent space alignment within the target. 
% Figure~\ref{fig:Method} overviews our structure-aware non-autoregressive neural machine translation framework. 
Figure~\ref{fig:Method} gives an overview of the network structure of our proposed \snat.
In \snat, the input sequence is segmented into sub-words by byte-pair tokenizer~\cite{sennrich2016neural}. In parallel, words in the input sequence are passed to POS and NER annotators to extract explicit syntactic and semantic structures, and the corresponding embeddings are aggregated by a linear layer to form the final syntax and semantic structure-aware embedding. The \snat model copies the structured encoder input as the decoder input and generates the translated sentences and labels. 

One of the most important properties of \snat is that it naturally introduces syntactic and semantic information when taking the structure-aware information as inputs and generating both words and labels. More precisely, given a source sentence $\mathbf{x}$, source label sequence $\mathbf{L_{x}}$, the conditional probability of a target translation $\mathbf{y}$ and target label sequence $\mathbf{L_{y}}$ is:%= \{L_{x_{1}}, L_{x_{2}}, \ldots, L_{x_{n}}\} = \{L_{y_{1}}, L_{y_{2}}, \ldots, L_{y_{m}}\}
\begin{equation}
\small{
P_{\rm \snat}(\mathbf{y}, \mathbf{L_y} | \mathbf{x}, \mathbf{L_{x}}; \varphi)= \prod_{t=1}^{m} p\left(y_{t}, L_{y_{t}}|  \hat{\mathbf{x}}, \hat{\mathbf{L}}_{\mathbf{x}}, \mathbf{x}, \mathbf{L_{x}} ; \varphi\right),
}
\end{equation}
where $\mathbf{x}$ and $\mathbf{L_x}$ are first fed into the encoder of \snat model. $\hat{\mathbf{x}}$ and $\mathbf{\hat{L}_{x}}$ with length $m$ are syntactic and semantic structure-aware copying of word and label from encoder inputs, respectively. We show the details in the following sections. 
\subsection{Syntactic and Semantic Labeling}
We use POS and NER to introduce the syntactic and semantic information existing in natural language, respectively. During the data pre-processing, each sentence is annotated into a semantic sequence using an open-source pre-trained semantic annotator. In particular, we take the Treebank style~\cite{marcus1999treebank} for POS and PropBank style~\cite{palmer2005proposition} for NER to annotate every token of input sequence with semantic labels. Given a specific sentence, there would be predicate-argument structures. Since the input sequence is segmented into subword units using byte-pair encoding~\cite{sennrich2016neural}, we assign the same label to all subwords tokenized from the same word. As shown in Figure \ref{fig:Method}, the word ``\textit{Ancelotti}'' is tokenized as ``\textit{An@@}'' and ``\textit{Celotti}''. The corresponding POS tags are PRON and PRON while the corresponding NER tags are B$\_$PERSON and I$\_$PERSON. For the text ``\textit{Is An@@ Celotti the man for the job ?}'', the corresponding POS tag set is \{AUX, PRON, PRON, DET, NOUN, ADP, DET, NOUN, PUNCT\} and the NER tag set is \{O, B$\_$PERSON, I$\_$PERSON, O, O, O, O, O, O\}. The data flow of the proposed model is also shown in Figure~\ref{fig:Method}.

\subsection{Encoder}
Following Transformer~\cite{vaswani2017attention}, we use a stack of 6 identical Transformer blocks as encoder. In addition to the word embedding and position embedding in the traditional Transformer, we add structure-aware label embedding. The input to the encoder block is the addition of the normalized word, labels (NER and POS) and position embedding, which is represented as $\mathbf{H}^{0}=[\mathbf{h}_{1}^{0}, \ldots, \mathbf{h}_{n}^{0}]$. 

The input representation $\mathbf{H}^{0}=[\mathbf{h}_{1}^{0}, \ldots, \mathbf{h}_{n}^{0}]$ is encoded into contextual layer representations through the Transformer blocks. For each layer, the layer representation $\mathbf{H}^{l}=[\mathbf{h}_{1}^{l}, \ldots, \mathbf{h}_{n}^{l}]$ is computed with the $l$-th layer Transformer block $\mathbf{H}^{l} = \mathrm{ Transformer}_{l}(\mathbf{H}^{l-1})$, $l \in \{1,2,\ldots,6\}$. In each Transformer block, multiple self-attention heads are used to aggregate the output vectors of the previous layer. 
A general attention mechanism can be formulated as the weighted sum of the value vector $\mathbf{V}$ using the query vector $\mathbf{Q}$ and the key vector $\mathbf{K}$:
\begin{equation}
\small
    \operatorname{Attention}(\mathbf{Q}, \mathbf{K}, \mathbf{V})=\operatorname{softmax}\left(\frac{\mathbf{Q} \mathbf{K}^{T}}{\sqrt{d_{\text {model}}}}\right) \cdot \mathbf{V},
\end{equation}
where $d_{\rm model}$ represents the dimension of hidden representations. For self-attention, $\mathbf{Q}$, $\mathbf{K}$, and $\mathbf{V}$ are mappings of previous hidden representation by different linear functions, i.e.,
% linearly projection of hidden representations of the previous layer, 
$\mathbf{Q} =\mathbf{H}^{l-1} \mathbf{W}_{Q}^{l}$, $\mathbf{K} =\mathbf{H}^{l-1} \mathbf{W}_{K}^{l}$, and  $\mathbf{V} =\mathbf{H}^{l-1} \mathbf{W}_{V}^{l}$, respectively. At last, the encoder produces a final contextual representation $\mathbf{H}^{6} = [\mathbf{h}^{6}_{1}, \ldots, \mathbf{h}^{6}_{n}]$, which is obtained from the last Transformer block.%the last layer of the encoder.

\subsection{Decoder}
The decoder also consists of 6 identical Transformer blocks, but with several key differences from the encoder. More concretely, we denote the contextual representations in the $i$-th decoder layer is $\mathbf{Z}^{i}(1 \leq i \leq 6)$. The input to the decoder block as $\mathbf{Z}^{0} = [\mathbf{z}_{1}^{0}, \ldots, \mathbf{z}_{m}^{0}]$, which is produced by the addition of the word, labels (NER and POS) copying from encoder input and positional embedding. 

For the target side input $[\hat{\mathbf{x}}, \hat{\mathbf{L}}_{\mathbf{x}}]$, most previous works simply copied partial source sentence with the length ratio $\frac{n}{m}$ where $n$ refers to the source length and $m$ is the target length as the decoder input. More concretely, the decoder input $y_{i}$ at the $i$-th position is simply a copy of the $\lfloor \frac{n}{m} \times i \rfloor$th contextual representation, i.e., $x_{\lfloor \frac{n}{m} \times i \rfloor}$ from the encoder.
From our investigation, in most cases, the gap between source length and target length is relatively small (e.g. 2).
Therefore, it deletes or duplicates the copy of the last few tokens of the source. If the last token is meaningful, the deletion will neglect important information. Otherwise, if the last token is trivial, multiple copies will add noise to the model. 

Instead, we propose a syntactic and semantic structure-aware mapping method considering the POS and NER labels to construct the decoder inputs. Our model first picks out the informative words with NOUN and VERB POS tags, and the ones recognized as entities by the NER module. If the source length is longer than the target length, we retain all informative words, and randomly delete the rest of the words. On the other hand, if the source length is shorter than the target, we retain all words and randomly duplicate the informative words. The corresponding label of a word is also deleted or preserved. Moreover, by copying the similar structural words from the source, it can provide more information to the target input than just copying the source token, which is greatly different from the target token. The POS and NER labels of those structure-aware copied words from the source sentence are also copied as the decoder input. So by using the structure-aware mapping, we can assign $[\hat{\mathbf{x}}, \hat{\mathbf{L}}_{\mathbf{x}}]$ as decoder input.

For positional attention which aims to learn the local word orders within the sentence~\cite{gu2018non}, we set positional embedding~\cite{vaswani2017attention} as both $\mathbf{Q}$ and $\mathbf{K}$, and hidden representations of the previous layer as $\mathbf{V}$. 

For inter-attention, $\mathbf{Q}$ refers to hidden representations of the previous layer, whereas $\mathbf{K}$ and $\mathbf{V}$ are contextual vectors $\mathbf{H}^{6}$ from the encoder. We modify the attention mask so that it does not mask out the future tokens, and every token is dependent on both its preceding and succeeding tokens in every layer. Therefore, the generation of each token can use bi-directional attention. The position-wise Feed-Forward Network (FFN) is applied after multi-head attention in both encoder and decoder. It consists of two fully-connected layers and a layer normalization~\cite{ba2016layer}. The FFN takes $\mathbf{Z}^{6}$ as input and calculates the final representation $\mathbf{Z}^{f}$, which is used to predict the whole target sentence and label:
\vspace{-1em}
\begin{equation}
\small
    \begin{split}
          p\left(\mathbf{y}\mid \hat{\mathbf{x}}, \hat{\mathbf{L}}_{\mathbf{x}}, \mathbf{x}, \mathbf{L_{x}}\right)=f\left(\mathbf{Z}^{f} \mathbf{W}_{w}^{\top}+\mathbf{b}_{w}\right),
    \end{split}
        \label{word}
\end{equation}
\vspace{-2.0em}
\begin{equation}
\small
    \begin{split}
   q\left(\mathbf{L_{y}}\mid \hat{\mathbf{x}}, \hat{\mathbf{L}}_{\mathbf{x}}, \mathbf{x}, \mathbf{L_{x}}\right)=f\left(\mathbf{Z}^{f} \mathbf{W}_{l}^{\top}+\mathbf{b}_{l}\right), 
    \end{split}
        \label{label}
\end{equation}
where $f$ is a GeLU activation function \cite{hendrycks2016gaussian}. $\mathbf{W}_{w}$ and $\mathbf{W}_{l}$ are the token embedding and structural label embedding in the input representation, respectively. We use different FFNs for POS and NER labels. To avoid redundancy, we just use $q\left(\mathbf{L_{y}}\mid \hat{\mathbf{x}}, \hat{\mathbf{L}}_{\mathbf{x}}, \mathbf{x}, \mathbf{L_{x}}\right)$ to represent the two predicted label likelihood in general.   
% , $\mathbf{W}_{n}$
%  
\subsection{Training}
% \paragraph{Word Decoding}
We use ($\mathbf{x}$, $\mathbf{L_{x}}$, $\mathbf{y}^{*}$, $\mathbf{L}_{\mathbf{y}}^{*}$) to denote a training instance. To introduce the label information, our proposed \snat contains a discrete sequential latent variable $L_{y_{1:m}}$ with conditional posterior distribution $p(L_{y_{1:m}}|\hat{\mathbf{x}}, \hat{\mathbf{L}}_{\mathbf{x}}, \mathbf{x}, \mathbf{L_{x}} ; \varphi)$. It can be approximated using a proposal distribution $q(\mathbf{L_{y}}\mid\hat{\mathbf{x}}, \hat{\mathbf{L}}_{\mathbf{x}}, \mathbf{x}, \mathbf{L_{x}})$. The approximation also provides a variational bound for the maximum log-likelihood:
\begin{equation}
\small
\begin{split}\label{eq:var-bound-loss}
  &\log P_{\rm \snat} = \log \sum_{t=1}^{m} q\left(L_{y_{t}}|  \hat{\mathbf{x}}, \hat{\mathbf{L}}_{\mathbf{x}}, \mathbf{x}, \mathbf{L_{x}} ; \varphi\right)\\ 
  &\times  p\left(y_{t}| L_{y_{t}}, \hat{\mathbf{x}}, \hat{\mathbf{L}}_{\mathbf{x}}, \mathbf{x}, \mathbf{L_{x}} ; \varphi\right) \\ 
 &\geq \underset{L_{y_{1: m}} \sim q}{\mathbb{E}} \Big\{ \underbrace{\sum_{t=1}^{m} \log q\left(L_{y_{t}} \mid \hat{\mathbf{x}},  \hat{\mathbf{L}}_{\mathbf{x}}, \mathbf{x}, \mathbf{L_{x}} ; \varphi\right)}_{\text {Label likelihood}}\\
 &+\underbrace{\sum_{t=1}^{m} \log p\left(y_{t} \mid L_{y_{t}}, \hat{\mathbf{x}}, \hat{\mathbf{L}}_{\mathbf{x}}, \mathbf{x}, \mathbf{L_{x}} ; \varphi\right)}_{\text {Structure-aware word likelihood}} \Big\} +\mathcal{H}(q). 
\end{split}
\end{equation}
Note that, the resulting likelihood function, consisting of the two bracketed terms in Eq.~(\ref{eq:var-bound-loss}), allows us to train the entire model in a supervised fashion. The inferred label can be utilized to train the label predicting model $q$ and simultaneously supervise the structure-aware word model $p$. 
The label loss can be calculated by the cross-entropy $\mathcal{H}$ of $L_{y_t}^{*}$ and Eq.~(\ref{label}):
\begin{equation}
\small
    \mathcal{L}_{\mathrm{label}}= \sum_{t=1}^{m} \mathcal{H}\left(L_{y_{t}}^{*}, ~~ q( L_{y_{t}} \mid \hat{\mathbf{x}}, \hat{\mathbf{L}}_{\mathbf{x}}, \mathbf{x}, \mathbf{L_{x}})\right), 
\end{equation}
The structure-aware word likelihood is conditioned on the generation result of the label. Since the Eq.~(\ref{word}) does not depend on the predicted label, we propose to bring the structure-aware word mask $\mathbf{M}_{wl} \in \mathbb{R}^{|V_{word}|\times |V_{label}|}$, where $|V_{word}|$ and $|V_{label}|$ are vocabulary sizes of word and label, respectively. The mask $\mathbf{M}_{w_l}$ is defined as follows:
\begin{equation}
\small
    \mathbf{M}_{w_l}(i,j) = \left\{  
    \begin{array}{lr}  
             1, ~~\mathcal{A}(y_{i}) =label_{j}, \\  
             \epsilon, ~~\mathcal{A}(y_{i}) \neq label_{j}, \\  
    \end{array}  
    \right. 
\end{equation}
% The mask $\mathbf{M}_{w_l}$ 
which can be obtained at the preprocessing stage, and $\mathcal{A}$ denotes the open-source pre-trained POS or NER annotator mentioned above.
It aims to penalize the case when the word $y_i$ does not belong to the label $label_j$ with $\epsilon$, which is a small number defined within the range of $(0,1)$ and will be tuned in our experiments. For example, the word ``great'' does not belong to VERB. The structure-aware word likelihood can be reformulated as:
\begin{equation}
% \footnotesize
\small
\begin{split}
    &p\left(y_{t} \mid L_{y_{t}}, \hat{\mathbf{x}}, \hat{\mathbf{L}}_{\mathbf{x}}, \mathbf{x}, \mathbf{L_{x}} ; \varphi\right) = \\&p(y_{t}\mid \hat{\mathbf{x}}, \hat{\mathbf{L}}_{\mathbf{x}}, \mathbf{x}, \mathbf{L_{x}}) \times \mathbf{M}_{wl} \times q(L_{y_{t}}\mid \hat{\mathbf{x}}, \hat{\mathbf{L}}_{\mathbf{x}}, \mathbf{x}, \mathbf{L_{x}}). 
\end{split}
\end{equation}
Consequently, the structure-aware word loss $\mathcal{L}_{\rm word}$ is defined as the cross-entropy between true $p^{'}(y_{t}^{*}| L_{y_t}^*)$ and predicted $p\left(y_{t} \mid L_{y_{t}}, \hat{\mathbf{x}}, \hat{\mathbf{L}}_{\mathbf{x}}, \mathbf{x}, \mathbf{L_{x}} ; \varphi\right)$,
where $p^{'}(y_{t}^{*}| L_{y_t}^*) \in \mathbb{R}^{|V_{word}|\times |V_{label}|}$ is a matrix where only item at the index of $(y_{t}^{*}, L_{y_t}^*)$ equals to 1, otherwise equals to 0. We reshape $p^{'}(y_{t}^{*}| L_{y_t}^*)$ and $p(y_{t}| L_{y_t})$ to vectors when calculating the loss.   
\paragraph{Intermediate Alignment Regularization}
One main problem of NAT is that the decoder generation process does not depend on the previously generated tokens. Based on the bidirectional nature of \snat decoder, the token can depend on every token of the decoder input. However, since the input of decoder $[\hat{\mathbf{x}}, \hat{\mathbf{L}}_{\mathbf{x}}]$ is the duplicate of encoder input $[\mathbf{x}, \mathbf{L_{x}}]$, the generation depends on the encoder tokens rather than the target $\mathbf{y}^{*}$. 

To solve this problem, we align the output of the intermediate layer of the decoder with the target. The alignment makes the generation of following layers dependent on the coarse target-side information instead of the mere encoder input. This alignment idea is inspired by~\cite{guo2019non}, which directly feeds target-side tokens as inputs of the decoder by linearly mapping the source token embeddings to target embeddings. However, using one FFN layer to map different languages to the same space can hardly provide promising results. Thus, instead of aligning the input of decoder with the target, we use the intermediate layer of decoder to align with the target. In this case, our model avoids adding additional training parameters and manages to train the alignment together with \snat model in an end-to-end fashion. Formally, we define the intermediate alignment regularization as cross-entropy loss between the predicted word and the true word:
\begin{equation}
    \small
       \mathcal{L_{\mathrm{reg}}}= \sum_{t=1}^{m} \mathcal{H}\left(y_{t}^*, ~~ \mathrm{FFN}(\mathbf{Z}_{t}^{md})\right),
\end{equation}
% It is easy to implement and no extra parameter is introduced to the training process.
where $\mathbf{Z}^{md}\ (1<md<6)$ represents the output of each intermediate layer. Consequently, the final loss of \snat can be represented with the coefficient $\lambda$ as: 
\begin{equation}
\small
    \mathcal{L_{\snat}} = \mathcal{L}_{\mathrm{word}} + \mathcal{L}_{\mathrm{label}} + \lambda\mathcal{L_{\mathrm{reg}}}.
\end{equation}

% To be concrete, given the output of the intermediate layer $\mathbf{Z}^{md}$, where $1<md<6$. 
% The intermediate alignment regularizer and final loss can be represented as: 

%% file: 04_EXP.tex
\section{Experiment}
\subsection{Experimental Setup}
\paragraph{Data}
We evaluate \snat performance on both the WMT14 En-De (around 4.5M sentence pairs) and the WMT16 En-Ro (around 610k sentence pairs) parallel corpora. For the parallel data, we use the processed data from~\cite{ghazvininejad2019mask} to be consistent with previous publications. The dataset is processed with Moses script~\cite{hoang2008design}, and the words are segmented into subword units using byte-pair encoding~\cite{sennrich2016neural}. The WMT14 En-De task uses newstest-2013 and newstest-2014 as development and test sets, and WMT16 En-Ro task uses newsdev-2016 and newstest-2016 as development and test sets. We report all results on test sets. The vocabulary is shared between source and target languages and has $\sim$36k units and $\sim$34k units in WMT14 En-De and WMT16 En-Ro, respectively. 

\begin{table*}[!t]
\centering
\caption{Performance of BLEU score on WMT14 En$\leftrightarrow$De and WMT16 En$\leftrightarrow$Ro tasks. ``-'' denotes that the results are not reported. LSTM-based results are from \cite{wu2016google}; CNN-based results are from \cite{gehring2017convolutional}. $^\dagger$The Transformer \cite{vaswani2017attention} results are based on our own reproduction.}
\resizebox{0.93\textwidth}{!}{
\begin{tabular}{lcccccc}
\toprule[1.5pt]
% & \multicolumn{2}{c}{\textbf{WMT14}} & \multicolumn{2}{c}{\textbf{WMT16}} &         &         \\ % \hline
& \textbf{En$\to$De} & \textbf{De$\to$En} & \textbf{En$\to$Ro} & \textbf{Ro$\to$En} &         &         \\ \hline
\textbf{\textit{Autoregressive Models}}                                       &                  &                 &                 &                   & \textbf{Latency} & \textbf{Speedup} \\ \hline
\textbf{LSTM Seq2Seq} \cite{bahdanau2016actor}      & 24.60             & -               & -               & -             & -       & -       \\
\textbf{Conv S2S} \cite{gehring2017convolutional}   & 26.43            & -               & 30.02               & -             & -       & -       \\
\textbf{Transformer$^\dagger$}   \cite{vaswani2017attention}                                              & 27.48            & 31.29           &    34.36             & 33.82             & 642ms  & 1.00X   \\ \hline
\textbf{\textit{Non-autoregressive Models}}                                   &                  &                 &                 &                   & \textbf{Latency} & \textbf{Speedup} \\ \hline
\textbf{NAT} \cite{gu2018non}                       & 17.69            & 20.62           & 29.79               & -                 & 39ms    & 15.6X   \\
\textbf{NAT, rescoring 10} \cite{gu2018non}                       & 18.66            & 22.41           & -               & -                 & 79ms    & 7.68X   \\
\textbf{NAT, rescoring 100} \cite{gu2018non}    & 19.17            & 23.20           & -               & -                 & 257ms    & 2.36X   \\
\textbf{iNAT} \cite{lee2018deterministic}           & 21.54            & 25.43           & 29.32               & -                 & -       & 5.78X   \\
\textbf{Hint-NAT} \cite{li2019hint}                 & 21.11            & 25.24           & -               & -             & 26ms    & 23.36X   \\
\textbf{FlowSeq-base}      \cite{ma2019flowseq}     & 21.45            & 26.16           & -               & 29.34             & -       & -       \\
% \textbf{CMLM} \cite{ghazvininejad2019mask}     & 18.05             &     21.83        &        24.23     &     -        &     -    &     -    \\ 
\textbf{ENAT-P} \cite{guo2019non}  & 20.26             &     23.23        &        29.85     &     -        &     25ms    &     24.3X    \\ 
\textbf{ENAT-P, rescoring 9}   & 23.22             &     26.67        &        34.04     &     -        &     50ms    &     12.1X    \\ 
\textbf{ENAT-E}   & 20.65             &     23.02        &        30.08     &     -        &     24ms    &     25.3X    \\ 
\textbf{ENAT-E, rescoring 19}  & 24.28             &     26.10        &       \underline{34.51}     &     -        &     49ms    &     12.4X    \\ 
\textbf{DCRF-NAT}    \cite{sun2019fast}             & 23.44            & 27.22           & -               & -             & 37ms    & 16.4X   \\
\textbf{DCRF-NAT, rescoring 9}              & 26.07            & 29.68           & -               & -             & 63ms    & 6.1X   \\
\textbf{DCRF-NAT, rescoring 19}             & \underline{26.80}            & \underline{30.04}           & -               & -             & 88ms    & 4.4X   \\
\textbf{NAR-MT(rescoring 11)}   \cite{zhou2020improving}          & 23.57           & 29.01           & 31.21               & 32.06                 & -       & -       \\
\textbf{NAR-MT(rescoring 11) + monolingual}            & 25.53           & 29.96           & 31.91               & \underline{33.46}                 & -       & -       \\
\textbf{LAVA NAT}   \cite{li2020lava}              & 25.29            & 30.31           & -               & 30.73             & 28ms   & 21.4X  \\
\textbf{AXE CMLM}   \cite{ghazvininejad2020aligned} & 23.53            & 27.90           & 30.75           & 31.54                 & -       & -       \\ \hline
\textbf{\snat}                          &    24.64          &   28.42          &      32.87       &     32.21        &  26.88ms       &   22.6X      \\
\textbf{\snat, rescoring 9}    &    26.87          &     30.12        &      34.93       &     33.11        &    54.63ms     &    11.1X     \\
\textbf{\snat, rescoring 19}     &     \textbf{27.50}         &    \textbf{30.82}     &    \textbf{35.19}         &  \textbf{33.98}           &    65.62ms     &    9.3X     \\

\bottomrule[1.5pt]
\end{tabular}
}
\label{result}
\end{table*}

\paragraph{Model Configuration}
Our implementation is based on the PyTorch sequence modeling toolkit Fairseq\footnote{\url{https://github.com/pytorch/fairseq}}. We follow the weights initialization scheme from BERT and follow the settings of the base Transformer configuration in~\cite{vaswani2017attention} for all the models: 6 layers per stack, 8 attention heads per layer, 512 model dimensions and 2,048 hidden dimensions. The dimension of POS and NER embedding is set to 512 which is the same as the word embedding dimension. The 
autoregressive and non-autoregressive model have the same encoder-decoder structure, except for the decoder attention mask and the decoding input for the non-autoregressive model as described in Sec.~\ref{sec:meth}. We try different values for the label mismatch penalty $\epsilon$ from $\{0.01, 0.1, 0.5\}$ and find that $0.1$ gives the best performance. The coefficient $\lambda$ is tested with different values from $\{0.25, 0.5, 0.75, 1\}$, and $\lambda=0.75$ outperforms other settings. We set the initial learning rate as values from $\{$8e-6, 1e-5, 2e-5, 3e-5$\}$, with a warm-up rate of 0.1 and L2 weight decay of 0.01. Sentences are tokenized and the maximum number of tokens in each step is set to 8,000. The maximum iteration steps is set to 30,000, and we train the model with early stopping.

\paragraph{Baselines} We choose the following models as baselines:
\textbf{NAT} is a vanilla non-autoregressive Transformer model for NMT which is first introduced in~\cite{gu2018non}.
\textbf{iNAT}~\cite{lee2018deterministic} extends the vanilla NAT model by iteratively reading and refining the translation. The number of iterations is set to 10 for decoding.
\textbf{Hint-NAT}~\cite{li2019hint} utilizes the intermediate hidden states from an autoregressive teacher to improve the NAT model.
\textbf{FlowSeq}~\cite{ma2019flowseq} adopts normalizing flows \cite{kingma2018glow} as latent variables for generation.
% \textbf{CMLM}~\cite{ghazvininejad2019mask} uses a masked language modeling objective to train a model to predict any subset of the target words, conditioned on both the input text and a partially masked target translation.
\textbf{ENAT}~\cite{guo2019non} proposes two ways to enhance the decoder inputs to improve NAT models. The first one leverages a phrase table to translate source tokens to target tokens \textbf{ENAT-P}. The second one transforms source-side word embedding into target-side word embeddings \textbf{ENAT-E}. 
\textbf{DCRF-NAT}~\cite{sun2019fast} designs an approximation of CRF for NAT models and further uses a dynamic transition technique to model positional context in the CRF.
\textbf{NAR-MT}~\cite{zhou2020improving} uses a large number of source texts from monolingual corpora to generate additional teacher outputs for NAR-MT training.
\textbf{LAVA NAT}~\cite{li2020lava} uses look-around strategy and vocabulary attention to learn the long-term token dependencies inside the decoder.
\textbf{AXE CMLM}~\cite{ghazvininejad2020aligned} trains the conditional masked language models using a differentiable dynamic program to assign loss based on the best possible monotonic alignment between target tokens and model predictions.

\subsection{Training and Inference Details}
To obtain the part-of-speech and named entity labels, we use industrial-strength spaCy\footnote{{\url{https://spacy.io/usage/models}}} to acquire the label for English, German, and Romanian input. In our implementation, there are 17 labels for POS in total, which are ADJ (adjective), ADV (adverb), ADP (ad-position), AUX (auxiliary), CCONJ (coordinating conjunction), DET (determiner), INTJ (interjection), NOUN (noun), NUM (numeral), PART (particle), PRON (pronoun), PROPN (proper noun), PUNCT (punctuation), SCONJ (subordinating conjunction), SYM (symbol), VERB (verb), and X (other). The NER task is trained on OntoNotes v5.0 benchmark dataset~\cite{pradhan2013towards} using formatted BIO labels and defines 18 entity types: CARDINAL, DATE, EVENT, FAC, GPE, LANGUAGE, LAW, LOC, MONEY, NORP, ORDINAL, ORG, PERCENT, PERSON, PRODUCT, QUANTITY, TIME, and WORK$\_$OF$\_$ART. 
% ``''

\paragraph{Knowledge Distillation}
Similar to previous works on non-autoregressive translation~\cite{gu2018non,shu2020latent,ghazvininejad2019mask}, we use sequence-level knowledge distillation by training \snat on translations generated by a standard left-to-right Transformer model (i.e., Transformer-large for WMT14 EN$\to$DE, and Transformer-base for WMT16 EN$\to$RO). Specifically, we use scaling NMT~\cite{ott2018scaling} as the teacher model. We report the performance of standard autoregressive Transformer trained on distilled data for WMT14 EN$\to$DE and WMT16 EN$\to$RO. We average the last 5 checkpoints to obtain the final model. We train the model with cross-entropy loss and label smoothing ($\epsilon = 0.1$).

\paragraph{Inference}
During training, we do not need to predict the target length $m$ since the target sentence is given. During inference, we use a simple method to select the target length for \snat~\cite{wang2019non,li2019hint}. First, we set the target length to $m^{\prime}=n+C$, where $n$ is the length of the source sentence and $C$ is a constant bias term estimated from the overall length statistics of the training data. Then, we create a list of candidate target lengths with a range of $[m^{\prime}-B, m^{\prime}+B]$ where $B$ is the half-width of the interval. Finally, the model picks the best one from the generated $2B+1$ candidate sentences. 
In our experiment, we set the constant bias term $C$ to 2 for WMT 14 EN$\to$DE, -2 for WMT 14 DE$\to$EN, 3 for WMT 16 EN$\to$RO, and -3 for WMT 14 RO$\to$EN according to the average lengths of different languages in the training sets. We set $B$ to 4 or 9, and obtain corresponding 9 or 19 candidate translations for each sentence. Then we employ an autoregressive teacher model to rescore these candidates. 

% For example, if $n=5, C=1$ and we use a half-width of $B=2$, the model will generate 5 candidate translations with different length 4,5,6,7, and 8. The autoregressive teacher model will rank 5 candidates, and select the one with the highest probability. This is referred to as the length-parallel decoding, as discussed in~\cite{wei2019imitation}.

% \begin{figure*}[t]
% \centering
% \includegraphics[width=1\linewidth]{table2.pdf}
% \end{figure*}

\subsection{Results and Analysis}
Experimental results are shown in Table~\ref{result}. We first compare the proposed method against autoregressive counterparts in terms of translation quality, which is measured by BLEU~\cite{papineni2002bleu}. For all our tasks, we obtain results comparable with the Transformer, the state-of-the-art autoregressive model. Our best model achieves 27.50 (+0.02 gain over Transformer), 30.82 (-0.46 gap with Transformer), 35.19 (+0.82 gain), and 33.98 (+0.16 gain) BLEU score on WMT14 En$\leftrightarrow$De and WMT16 EN$\leftrightarrow$Ro, respectively. More importantly, our \snat decodes much faster than the Transformer, which is a big improvement regarding the speed-accuracy trade-off in AT and NAT models.

Comparing our models with other NAT models, we observe that the best \snat model achieves a significant performance boost over NAT, iNAT, Hint-NAT, FlowSeq, ENAT, NAR-MT and AXE CMLM by +8.33, +5.96, +6.39, +6.05,  +3.22, 3.93 and +3.97 in BLEU on WMT14 En$\to$De, respectively. This indicates that the incorporation of the syntactic and semantic structure greatly helps reduce the impact of the multimodality problem and thus narrows the performance gap between Autoregressive Transformer (AT) and Non-Autoregressive Transformer (NAT) models. In addition, we see a +0.69, +0.78, +0.68, and 0.52 gain of BLEU score over the best baselines on WMT14 En$\to$De, WMT14 De$\to$En, WMT16 En$\to$Ro and WMT16 Ro$\to$En, respectively. 

From the result of our methods at the last group in Table~\ref{result}, we find that the rescoring technique substantially assists in improving the performance, and dynamic decoding significantly reduces the time spent on rescoring while further accelerating the decoding process. On En$\to$De, rescoring 9 candidates leads to a gain of +2.23 BLEU, and rescoring 19 candidates gives a +2.86 BLEU score increment. 

\paragraph{Decoding Speed}
Following previous works~\cite{gu2018non,lee2018deterministic,guo2019non}, we evaluate the average per-sentence decoding latency on WMT14 En$\to$De test sets with batch size being 1, under an environment of NVIDIA Titan RTX GPU for the Transformer model and the NAT models to measure the speedup. The latencies are obtained by taking an average of five runs. More clearly, We reproduce the Transformer on our machine. We copy the runtime of other models but the speedup ratio is between the runtime of their implemented Transformer and their implemented model. We think it’s reasonable to compare the speedup ratio because it is independent of the influence caused by different implemented software or machines.

We can see from Table~\ref{result} that the best \snat gets a 9.3 times decoding speedup than the Transformer, while achieving comparable or even better performance. Compared to other NAT models, we observe that the \snat model is almost the fastest (only a little bit behind of ENAT and Hint-NAT) in terms of latency, and is surprisingly faster than DCRF-NAT with better performance. 

% \begin{figure*}[t]
% \centering
% \includegraphics[width=0.89\linewidth]{table3.pdf}
% \end{figure*}

% \begin{figure*}[t]
% \centering
% \includegraphics[width=0.89\linewidth]{table5.pdf}
% \end{figure*}
\begin{table}[ht]
\caption{The performance of different vision of SNAT models on WMT14 En$\to$De development set. \ding{52} means selecting the label tag.}
\centering
\resizebox{0.38\textwidth}{!}{
\begin{tabular}{l|cc|c}
\cline{2-2}
\toprule[1.5pt]
\textbf{Model} & \textbf{POS tag} & \textbf{NER tag} & ~~ \textbf{BLEU} \\ \hline
% \textbf{SNAT}      &    \ding{52}   &    \ding{52}      &  ~~24.64   \\ \hline
SNAT-V1      &   \ding{52}   &      & ~~24.21 \\ \hline 
SNAT-V2      &       &  \ding{52}     & ~~24.09  \\ \hline
SNAT-V3      &       &               & ~~22.84  \\ 
\bottomrule[1.5pt]
\end{tabular}
}
\label{ablation}
\end{table}

\begin{table}[ht]
\centering
\caption{The performance with respect to using different layer of intermediate interaction. Evaluated by the BLEU score on WMT14 En$\to$De$|$WMT14 De$\to$En.}
\resizebox{0.45\textwidth}{!}{
\begin{tabular}{lcccc}
\toprule[1.5pt]
\textbf{Method} & \textbf{WMT14 En}$\to$ \textbf{De} & \textbf{WMT14 De}$\to$ \textbf{En} \\ \hline
w/o  &            23.11                  &           27.03                   \\
w/ $\textbf{Z}^2$   &       24.32           &           28.21                   \\
w/ $\textbf{Z}^3$   &       24.57           &           28.42                   \\ 
\bottomrule[1.5pt]
\end{tabular}
}
\label{layer}
\end{table}

\begin{table}[ht]
\centering
\caption{The performance with respect to different sentence lengths. Evaluated by the BLEU score on WMT14 En$\to$De.}
\resizebox{0.48\textwidth}{!}{
\begin{tabular}{lccccc}
\toprule[1.5pt]
\textbf{Model} & ~~~~\textbf{10}&~~~~\textbf{20} & ~~~~\textbf{30}&~~~~ \textbf{50}&~~~~ \textbf{100} \\ \hline
AT   &        ~~ 28.35        & ~~    28.32           &   ~~   28.30          &  ~~   24.26  &  ~~   20.73         \\ 
NAT &         ~~ 21.31              &    ~~ 19.55   &         ~~ 17.19              &    ~~ 16.31  &         ~~ 11.35                               \\
SNAT  &        ~~ 28.67      &   ~~ 28.50    &   ~~ 27.33 &   ~~ 25.41 &   ~~17.69    \\
\bottomrule[1.5pt]
\end{tabular}
}
\label{length}
\end{table}

\subsection{Ablation Analysis}
\paragraph{Effect of Syntactic and Semantic Structure Information} We investigate the effect of using the syntactic and semantic tag on the model performance. Experimental results are shown in Table~\ref{ablation}. It demonstrates that incorporating POS information boosts the translating performance (+1.37 on WMT14 En$\to$De) and NER information can also enhance the translating performance (+1.25 on WMT14 En$\to$De). The POS label enriches the model with the syntactic structure, while the NER label supplements the semantic information to the model which are critical elements for \snat model to exhibit better translation performance.

\paragraph{Effect of Intermediate Representation Alignment}
We conduct experiments for our \snat model on WMT14 En$\to$De with various alignments between decoder layers and target. As shown in Table~\ref{layer}, using the second layer $\mathbf{Z}^{2}$ in the decoder as intermediate alignment can gain +1.21 improvement, while using the third layer $\mathbf{Z}^{3}$ in the decoder as intermediate alignment can gain +1.46 improvement. This is in line with our expectation that aggregating layer-wise token information in intermediate layers can help improve the decoder’s ability to capture token-token dependencies.

\paragraph{Effect of Sentence Length}
To evaluate different models on different sentence lengths, we conduct experiments on the WMT14 En$\to$De development set and divide the sentence pairs into different length buckets according to the length of the reference sentences. For example, in Table~\ref{length}, the column of 100 calculates the BLEU score of sentences that the length of the reference sentence is larger than 50 but smaller or equal to 100. We can see that the performance of vanilla NAT drops quickly as the sentence length increases from 10 to 50, while AT model and the proposed \snat model have relatively stable performance over different sentence lengths. This result confirms the power of the proposed model in modeling long-term token dependencies.